\documentclass{article}

\usepackage{emnlp2020}

\usepackage[utf8]{inputenc} 
\usepackage[T1]{fontenc}    
\usepackage{hyperref}       
\usepackage{url}            
\usepackage{booktabs}       
\usepackage{amsfonts}       
\usepackage{nicefrac}       
\usepackage{microtype}      
\usepackage{xcolor,colortbl}
\usepackage{graphicx}
\usepackage{multirow}
\usepackage{background}     
\usepackage{placeins}
\usepackage{ifthen}
\usepackage[detect-weight=true, detect-family=true]{siunitx}
\usepackage{bm}

\backgroundsetup{
  position=current page.north,
  angle=0,
  vshift=-5mm,
  opacity=1,
  scale=1,
  contents=
}

\aclfinalcopy
\newcommand{\round}[1]{\num[round-mode=places,round-precision=1]{#1}}

\newcommand{\acc}[2]{\round{#1}&\ifthenelse{\equal{#2}{}}{}{\tiny ${\scriptstyle \pm}$\round{#2}}}

\newcommand{\graycellcolor}{\cellcolor{gray!25}}

\newcommand{\gacc}[2]{\graycellcolor\round{#1}&\graycellcolor\ifthenelse{\equal{#2}{}}{}{\tiny ${\scriptstyle \pm}$\round{#2}}}

\newcommand{\ACC}[2]{\textbf{\round{#1}}&\ifthenelse{\equal{#2}{}}{}{\textbf{\tiny ${\scriptstyle \pm}$\round{#2}}}}

\newcommand{\GACC}[2]{\graycellcolor\textbf{\round{#1}}&\graycellcolor\ifthenelse{\equal{#2}{}}{}{\textbf{\tiny ${\scriptstyle \pm}$\round{#2}}}}

\newcommand{\none}{-&}

\newcommand{\mc}[2]{\multicolumn{#1}{c}{\textbf{#2}}}

\title{Compositional Generalization in Semantic Parsing: \\ Pre-training vs. Specialized Architectures}

\date{}

\author{%
\bf Daniel Furrer\thanks{* Equal contribution. A description of each author’s contribution is available in Appendix~\ref{sect:contributions}.}
\hspace{0.2cm}
Marc van Zee\setcounter{footnote}{0}\footnotemark{}
\hspace{0.2cm}
Nathan Scales
\hspace{0.2cm}
Nathanael Sch\"arli\\
Google Research, Brain Team \\\\
\texttt{\{danielfurrer,marcvanzee,nkscales,schaerli\}@google.com}
}

\begin{document}

\maketitle

\begin{abstract}
While mainstream machine learning methods are known to have limited ability to compositionally generalize, new architectures and techniques continue to be proposed to address this limitation. We investigate state-of-the-art techniques and architectures in order to assess their effectiveness in improving compositional generalization in semantic parsing tasks based on the SCAN and CFQ datasets.
We show that masked language model (MLM) pre-training rivals SCAN-inspired architectures on primitive holdout splits. On a more complex compositional task, we show that pre-training leads to significant improvements in performance vs. comparable non-pre-trained models, whereas architectures proposed to encourage compositional generalization on SCAN or in the area of algorithm learning fail to lead to significant improvements.
We establish a new state of the art on the CFQ compositional generalization benchmark using MLM pre-training together with an intermediate representation.
\end{abstract}

\section{Introduction}


Human intelligence exhibits {\em systematic compositionality} \citep{fodor1988compositionality}, the capacity to understand and produce a potentially infinite number of novel combinations of  known components, i.e., to make ``infinite use of  finite means''~\citep{chomsky1965}. Humans demonstrate this ability across diverse domains, including natural language understanding (NLU)~\citep{Lake2018GeneralizationWS} and visual scene understanding~\citep{CLEVR,higgins2017scan}. For example, we can learn the meaning of a new word and then apply it to other language contexts. As \citet{Lake2018GeneralizationWS} put it: ``Once a person learns the meaning of a new verb `dax', he or she can immediately understand the meaning of `dax twice' and `sing and dax'.''


In contrast, state-of-the-art machine learning (ML) methods often fail to capture the compositional structure that is underlying the problem domain and thus fail to generalize compositionally \citep{Keysers2020,Lake2018GeneralizationWS,bastings2018jump,rearranging,CLEVR}.

The past several years have seen an increased focus on compositional generalization in both natural language and other domains, leading to the creation of datasets and specialized train-test splits to test compositional generalization~\citep{Lake2018GeneralizationWS,finegan2018text2sql,Keysers2020,CLEVR,hudson2019nsm,sinha2019clutrr} and the development of specialized architectures and training techniques intended to enforce a bias toward compositional generalization~\citep{li2019compositional,Gordon2020Permutation,nye2020learning,hudson2019nsm,neuralgpu,nsen,andreas2019good,hill2019environmental,lake2019compositional}.

In the area of natural language -- and semantic parsing in particular -- SCAN~\citep{Lake2018GeneralizationWS} has become a de facto standard benchmark and test bed for research in compositional generalization, inspiring a series of new techniques
~\citep{li2019compositional,russin2019comp,andreas2019good,lake2019compositional,Gordon2020Permutation,nye2020learning}. These techniques indeed lead to significant gains over vanilla baselines, including up to perfect performance on the primitive holdout splits. However, they frequently involve tailoring to the SCAN task, which raises the question to what degree these improvements would translate to other tasks.

Recently, \citet{Keysers2020} introduced the Compositional Freebase Questions (CFQ), a more realistic and larger scale semantic parsing dataset with rich compositional structure, together with the Distribution Based Compositionality Assessment (DBCA) method, which seeks to measure compositional generalization more comprehensively by making the distributions of compounds as different as possible between train and test, while keeping the distribution of atoms the same. While baseline evaluations in \citet{Keysers2020} show that three sequence-to-sequence architectures all struggle with compositional generalization on DBCA-based splits of both SCAN and CFQ, techniques designed specifically for compositional generalization on SCAN or other tasks have yet to be evaluated with DBCA.

At the same time, significant gains have been achieved in recent years in natural language tasks including semantic parsing through the use of MLM pre-training with systems such as BERT~\citep{devlin2019bert} and T5~\citep{raffel2019exploring} and in semantic parsing in particular through the introduction of intermediate representations that simplify the output format or map more closely to the input syntax~\citep{guo2019complex}. Neither of these techniques has yet been evaluated, however, in semantic parsing scenarios specifically designed to test compositional generalization.

In this paper, we address some of those open questions and make the following contributions:
\begin{itemize}
    \item We provide an overview of a range of architectures and techniques that have previously been applied to SCAN or CFQ and provide the most comprehensive summary of results so far (Table~\ref{table:results}), including results on detailed SCAN splits of two architectures (Transformer, Universal Transformer) that were previously evaluated only on CFQ and SCAN MCD splits.
    
    \item Using the DBCA method on CFQ and SCAN, we evaluate two representative architectures that had been proposed to improve compositional generalization (CGPS, Neural Shuffle Exchange Network) and find that neither show significant improvement compared to general-purpose sequence-to-sequence architectures.

    \item We evaluate the use of MLM pre-training on SCAN and CFQ and demonstrate that off-the-shelf pre-training is sufficient to nearly match the gains on SCAN's ``Add jump'' primitive holdout split achieved to date by SCAN-inspired specialized techniques, while obtaining significant improvements over comparable non-pre-trained models on CFQ.
    
    
    \item By combining pre-training with an intermediate representation, we obtain a new state-of-the-art score for CFQ of \textbf{42.1\%} on the MCD-mean split, beating the previous best results of 18.9\%.
\end{itemize}

This paper is organized as follows: Section~\ref{sect:background} describes SCAN and CFQ, our two compositional generalization benchmarks of interest; Section~\ref{sect:architectures_and_techniques} provides and overview of all architectures and techniques, which are then evaluated and analyzed in Section~\ref{sect:results}, and Section~\ref{sect:related-work} discusses related work.

\section{Background: Measuring Compositional Generalization}
\label{sect:background}

\subsection{SCAN}
SCAN~\citep{Lake2018GeneralizationWS} is a semantic parsing dataset of natural language navigation commands (e.g. \verb|turn left twice and jump|) that are mapped to corresponding action sequences (e.g. \verb|LTURN LTURN JUMP|). The SCAN dataset is compositional in the sense that it is constructed from a simple grammar consisting of rules correponding to primitive commands such as \verb|jump| and \verb|turn left| as well as rules that compose actions into more complex sequences.

Based on this explicit construction, \citet{Lake2018GeneralizationWS} and \citet{rearranging} introduce various experiments using train/test splits that are specifically designed to measure compositional generalization. For example, some experiments perform the train/test split based on the length of the commands, while other experiments make sure that a certain primitive command (such as \verb|jump|) only occurs in very limited combinations during training. Yet other experiments hold out whole subcommands (such as \verb|jump around right|) or templates (such as \verb|$Primitive around left|) in training.

\subsection{DBCA and CFQ}
\label{sect:background:cfq}
While all SCAN experiments are insightful, it is not clear which experiments are most relevant for measuring compositional generalization \citep{bastings2018jump}, and the performance of individual architectures varies significantly across different experiments (see Table~\ref{table:results}).~\citet{Keysers2020} address this concern by introducing distribution-based compositionality assessment (DBCA), which is a novel method to construct compositionality benchmarks more systematically.

Given a dataset (like SCAN) that is generated from a set of rules, they track for each example the individual rules (atoms) and  rule combinations (compounds) that were used to construct the example. Using this information, they then generate ``maximum compound divergence'' (MCD) splits, which maximize the compound divergence while guaranteeing a small atom divergence between train and test sets. MCD splits are well suited for assessing compositionality because they are both fair (because the distribution of individual rules is similar) and compositionally challenging (because the distribution of compounds is as different as possible).

Together with the DBCA method, \citet{Keysers2020} also provide CFQ, which is a simple but realistic and large natural language dataset that is specifically designed to measure compositional generalization using DBCA.
The task of interest is semantic parsing from a natural language question (such as 'Which art director of [Stepping Sisters 1932] was a parent of [Imre Sándorházi]?') to a SPARQL query, which can then be executed against the Freebase knowledge base. Named entities are anonymized, which is standard practice and ensures that the models do not have to learn all the entities.

%

The authors release a number of MCD splits for both SCAN and CFQ, and they show that there is a strong negative correlation between the accuracy of three standard sequence-to-sequence architectures and the compound divergence. They investigate this for \textit{LSTM+attention} (an LSTM \citep{hochreiter1997long} with attention mechanism \citep{bahdanau-attention-iclr}), for \textit{Transformer} \citep{vaswani2017attention}, and for \textit{Universal Transformer} \citep{dehghani2018universal}.

\section{Architectures and Techniques}
\label{sect:architectures_and_techniques}

In this section we give an overview of general sequence-to-sequence architectures that have been applied in the past to the SCAN or CFQ benchmarks, together with specialized architectures and techniques (SCAN-inspired and algorithm learning-based approaches), and MLM pre-trained models. We present in Section~\ref{sect:results} the results that each of these approaches obtains.

\subsection{General Sequence-to-sequence Architectures}

\paragraph{LSTM} 
\citet{Lake2018GeneralizationWS} evaluate simple recurrent networks (SRNs; \citet{elman1990finding}), long short-term memory networks \citep{hochreiter1997long}, and gated recurrent units (GRUs; \citet{chung2014empirical}), each with and without an attentional mechanism \citep{bahdanau-attention-iclr} on a number of SCAN splits. They find the overall-best architecture (determined by hyperparameter search) to be an LSTM without attention. 

\paragraph{CNN} 
\citet{dessi2019cnns} evaluate convolutional networks on some SCAN splits and find that ``CNNs are dramatically better than RNNs at compositional generalization''.

\paragraph{GRU, GRU-dep} 
\citet{bastings2018jump} analyze SCAN and conclude that it has very few target-side dependencies, and therefore simpler architectures tend to perform better on it. They focus on RNNs with GRU cells and also evaluate a GRU-based architecture that is non-autogressive [GRU-dep] which indeed performs better. 

\paragraph{LSTM+A, Transformer, Universal Transformer}
\citet{Keysers2020} evaluate these three architectures using DCBA splits of CFQ and SCAN. Hyperparameters are obtained using a search on a random CFQ split.

\paragraph{Evolved Transformer} \citet{evolved_transformer} found this architecture via an evolutionary neural architecture search that was seeded with a standard Transformer architecture. The most notable difference from a standard Transformer is the use of wide depth-wise separable convolutions in the early layers of both the encoder and decoder. The evolution was done on an English-German translation task, and the final architecture consistently beats the standard Transformer in several translation tasks as well as language modeling.
\citet{yun2019transformers} independently show that depth-wise separable convolutions have interesting theoretical properties, being more efficient than attention while keeping the transformer architecture functionally universal.

\subsection{SCAN-inspired Approaches}
The SCAN dataset has inspired many models that improve compositional generalization in some ways. Most of these approaches are guided by the observation that the mapping of input to output surface forms (e.g. the command \verb|jump| to the action \verb|JUMP|) can be handled separately from the mapping of the structure (e.g. \verb|X twice after Y| to \verb|Y X X|). A model that decouples these concerns - i.e. is equivariant to substitution of primitives - generalizes better. 

\paragraph{Syn-att}
\citet{russin2019comp} call this decoupling ``separating syntax from semantics'' and draw on neurosience research to further motivate their approach. They implement two separate encodings for each input token: one which maps each token independently (without context) and one which uses recurrence. This latter representation is used as an attention map over the former.

\paragraph{CGPS}
\citet{li2019compositional} propose an almost identical architecture but add entropy regularization to the encodings and show that this is important for reducing variance. The model is thoroughly evaluated on many SCAN splits and performs well on most. Like Syn-att, it is also conceptually simple and does not require complicated learning setups or additional supervision (unlike many other models mentioned below).

We select this model as a representative of SCAN-inspired approaches to evaluate the degree to which its gains carry over to the MCD splits of SCAN and CFQ. For CFQ we adapt the setup slightly to deal with the fact that the architecture assumes that every output token is directly mappable from an input token: We add a prefix whose token length is equal to the number of tokens not otherwise directly mappable (in particular, SPARQL syntactic tokens such as \verb|SELECT|, etc.) to all inputs. We also run a hyperparameter search to optimize for the CFQ task (see Appendix \ref{suppl:hyperparameters}).

\paragraph{Equivariant}
\citet{Gordon2020Permutation} seek to achieve a similar equivariance across primitive substitutions, building on the conceptually appealing notion of group convolutions~\citep{cohen2016group,kondor2018generalization}. The applicability of this approach in its current form is limited by the need for task-specific pairs of related input/output forms to be explicitly provided as a side input. 

\paragraph{GECA}
\citet{andreas2019good} introduces a data augmentation technique called Good Enough Compositional Data Augmentation (GECA) that seeks to provide a compositional inductive bias by automatically detecting ``templates'' that occur multiple times in training and then mechanically constructing additional training examples from these templates by filling them with different text ``fragments'' that were observed in similar contexts to the original fragments. Despite the tendency to introduce grammatically or semantically ``incorrect'' examples, the authors report significant improvements on several SCAN splits. (See also \citet{ruis2020benchmark}, however, which reports largely negative results for GECA when applied to a different compositional task).

\paragraph{LANE}
Concurrently with this work,~\citet{liu2020compositional} propose a modular architecture with memory, which consists of two cooperating modules -- a Composer and a Solver -- trained with hierarchical reinforcement learning using a curriculum. The Composer obtains ``analytical expressions'' (which are sequences with words and variables) from the inputs, while the Solver converts a variable-based source expression into a corresponding variable-based destination expression and then assigns the variables through interaction with memory. This approach has the benefit of not depending on the sorts of task-specific extra resources required by the meta-learning approaches described below, while demonstrating perfect generalization on SCAN splits including length and MCD. We expect that the method for identifying and composing expressions would need modification in order to apply it to CFQ, which the authors propose to do in future work.

\subsubsection{Meta-learning for SCAN}

Another line of research seeks to improve performance on SCAN through meta-learning, in which a system ``learns to learn'' by being trained on batches of examples generated by grammars that are distinct from, but structurally related to, the true SCAN grammar that is used in evaluation. These meta-learning approaches suffer the disadvantage of requiring task-specific supplementary data at training time, in this case in the form of the meta-learning training data, whose generation requires access to (or manual construction of) a family of grammars that are known to be structurally close to the grammar to be learned.

\paragraph{Meta seq2seq}
\citet{lake2019compositional} introduces a meta-learning model which takes as an additional input a set of support input/output pairs. When trained with episodes of carefully crafted problems that are similar to the problem of interest the model learns to use the support set to aid generalization. For example, a model trained with episodes of SCAN samples where the command to action meaning was shuffled (e.g. \verb|jump| $\rightarrow$ \verb|LOOK|) can solve the SCAN ``Add jump`` task if provided with the true command to action mapping as support set.

\paragraph{Synth}
\citet{nye2020learning} propose a neural program synthesis approach, in which the system, when exposed to a carefully crafted meta-learning training regime, is able to learn to reverse engineer a symbolic grammar equivalent to the original SCAN grammar. They evaluate their approach on a variety of SCAN splits and achieve perfect scores including on the challenging length split.

\subsection{Algorithm Learning}

\paragraph{NSEN}
The Neural GPU \citep{neuralgpu,freivalds2017improving} achieves perfect generalization on a number of algorithmic tasks such as binary addition and multiplication. More recently, \citet{nsen} show how their Neural Shuffle-Exchange Network (NSEN) outperforms the Neural GPU in many ways and does well on algorithmic problems as well as a language task.
We select NSEN as a representative of algorithm learning approaches to evaluate the degree to which it can systematically learn the algorithms of SCAN and CFQ. Both of these tasks can be expected to be learnable algorithmically due to their rule-based nature.

\subsection{Masked Language Model Pre-training}
\label{sect:pretraining}

Pre-training large, deep neural language models and successive fine-tuning on a variety of downstream natural language processing (NLP) tasks has yielded impressive results ~\citep{devlin2019bert,yang2019xlnet,raffel2019exploring}. The general idea behind pre-training is that it provides the model with general knowledge of syntax and ``world knowledge''. While MLM pre-training is a common approach in some semantic parsing domains~\citep{hwang2019comprehensive,guo2019complex,choi2020ryansql,wang2019rat}, this has not yet been tried on SCAN or CFQ.


The Text-to-Text Transfer Transformer (T5)~\cite{raffel2019exploring} treats every NLP task as a text-to-text problem, and is therefore suitable for the semantic parsing tasks we consider in this paper.~\citet{raffel2019exploring} show state-of-the-art results on many NLP tasks such as SQuAD, MultiRC, and BoolQ. T5 is released with multiple pre-trained models, ranging from ``small'' (60 million parameters) to ``11B'' (11 billion parameters).

\paragraph{T5-small, T5-small-NP, T5-base, T5-large, T5-3B, T5-11B} In our experiments, we fine-tune all T5 variations on both SCAN and CFQ. For comparison we also train a T5-small model without loading the pre-trained weights (\textbf{T5-small-NP}). We use the T5 public pre-trained models\footnote{\url{https://console.cloud.google.com/storage/browser/t5-data/}} and fine-tune them with an inverse square root learning rate schedule on CFQ with $10^4$ warm-up steps\footnote{This is the same schedule that has been used to pre-train the T5 models.}, and a constant learning rate of 0.003 on SCAN (as opposed to the constant learning rate of 0.001 that is used for fine-tuning in the original T5 paper). We use a batch size of $2^{17}$ tokens per batch. 
In the original T5 paper, all models are fine-tuned for 256k steps, but we found this number to be too large for SCAN due to the small size of the dataset, which led to overfitting. In order to determine the number of steps to fine-tune, we fine-tune on the ``simple'' split and choose the number of steps when the model converges, which is around 20k. We replicate most runs 5 times and report the average accuracy and confidence intervals, except for the non-MCD splits where we run only 1 replica for models larger than T5-base. As the variance in the runs with 5 replicas was relatively small, we do not expect the use of single replicas on some splits to affect the overall results we present.

T5's vocabulary does not contain curly braces (see ~\citet{raffel2019exploring}; Section 2.2 for their rationale). To work around this, we relax the evaluation to accept an out-of-vocabulary token in the output wherever a curly brace is expected.

\subsection{Intermediate SPARQL Representation}
\label{sect:output-simplification}

\citet{guo2019complex} show that significant gains can be obtained on text-to-SQL semantic parsing tasks by using an intermediate representation to help address the mismatch between intents expressed in natural language and SQL. We observe that the structure of natural language questions in CFQ and their semantic parses as SPARQL query can be quite different and that ML models struggle with such examples. Consider for example the question
\begin{quote}
Did \underline{M0 and M1} direct \underline{M2 and M3}?
\end{quote}
which is interpreted with SPARQL clauses like
\begin{quote}
\texttt{M0 directed M2 . M1 directed M2 . M0 directed M3 . M1 directed M3}
\end{quote}
It is straightforward to translate this to a version that is structurally more aligned with the question
\begin{quote}
\texttt{\{M0, M1\} directed \{M2, M3\}}
\end{quote}
by preprocessing the example outputs for training and then post-processing to obtain the original valid SPARQL form at inference time.
We show that such an intermediate representation results in better performance on CFQ and provide more details in Appendix~\ref{appendix:sparql}.\\

\noindent
\textbf{T5-11B-mod}
We evaluate T5-11B on the CFQ MCD split using the intermediate SPARQL representation where we group both subjects and objects.

\section{Results and Analysis}
\label{sect:results}

\begin{table*}[ht!]
\centering
\resizebox{\textwidth}{!}{
\setlength{\tabcolsep}{1pt}
\begin{tabular}{lrlrlrlrlrlrlrlrlrl}
                &            &      & \mc{2}{Add}           & \mc{2}{Jump}       &   &                &     &              &       &           &       &            &        &        & &\\
                & \mc{2}{Add}       & \mc{2}{turn}          & \mc{2}{around}     & \mc{2}{Around}     & \mc{2}{Opposite}  &       &          &    &              & \mc{2}{SCAN}  &\mc{2}{CFQ}\\
\textbf{Model}  & \mc{2}{jump}      & \mc{2}{left}          & \mc{2}{right}      & \mc{2}{right}      & \mc{2}{right}     & \mc{2}{Right}    & \mc{2}{Length}    & \mc{2}{MCD}   & \mc{2}{MCD}\\
\hline
\hline
LSTM            & \acc{0.08}{}      & \acc{90.3}{}           & \acc{98.43}{0.54}  & \acc{2.46}{2.68}  & \acc{47.62}{17.72}& \acc{23.49}{8.09}& \acc{13.8}{}      & \none{}              & \none{}\\
LSTM+A          & \gacc{0.0}{0.0}   & \gacc{82.6}{8.2}       & \GACC{100.0}{0.0}  & \gacc{0.0}{0.0}   & \gacc{16.5}{6.4}  & \gacc{30.0}{7.8} & \gacc{14.1}{}     & \acc{6.1}{1.7} & \acc{14.9}{1.1}\\
CNN             & \acc{69.2}{9.2}   & \none{}                & \none{}            & \acc{56.7}{10.2}  & \none{}           & \none{}          & \acc{0.0}{}       & \none{}             & \none{}\\
GRU             & \acc{12.5}{6.6}   & \acc{59.1}{16.8}       & \none{}            & \none{}           & \none{}           & \none{}          & \acc{18.1}{}      & \none{}             & \none{}\\
GRU-dep         & \acc{0.7}{0.4}    & \acc{90.8}{3.6}        & \none{}            & \none{}           & \none{}           & \none{}          & \acc{17.8}{}      & \none{}             & \none{}\\
Transformer     & \gacc{1.0}{0.6}   & \GACC{99.6}{0.8}       & \GACC{100.0}{0.0}  & \gacc{53.3}{10.9} & \gacc{3.0}{6.8}   & \gacc{92.0}{15.1}& \gacc{0.0}{}      & \acc{0.9}{0.3}     & \acc{17.8}{0.9}\\
Univ. Trans.    & \gacc{0.3}{0.3}   & \GACC{99.4}{1.4}       & \GACC{100.0}{0.0}  & \gacc{47.0}{10.0} & \gacc{15.2}{13.0} & \gacc{83.2}{18.2}& \gacc{0.0}{}      & \acc{1.1}{0.6} & \acc{18.9}{1.4}\\
Evol. Trans.    & \gacc{0.6}{0.6}   & \GACC{100.0}{0.0}      & \GACC{100.0}{0.0}  & \gacc{30.2}{28.4} & \gacc{11.6}{14.6} & \GACC{99.9}{0.3} & \gacc{19.8}{0.0}  & \gacc{1.6}{0.6}& \gacc{20.8}{0.7}\\
\hline
Syn-att         & \acc{91.0}{27.4}  & \ACC{99.9}{0.16}       & \acc{98.9}{2.3}    & \acc{28.9}{34.8}  & \acc{10.5}{8.8}   & \acc{99.1}{1.8}  & \acc{15.2}{0.7}   & \none{}             & \none{}\\
CGPS            & \acc{98.8}{1.4}   & \ACC{99.7}{0.4}        & \ACC{100.0}{0.0}   & \acc{83.2}{13.2}  & \acc{89.3}{5.5}   & \ACC{99.7}{0.5}  & \acc{20.3}{1.1}   & \gacc{2.0}{0.7}& \gacc{7.1}{1.8}\\
Equivariant*    & \acc{99.1}{0.04}  & \none{}                & \none{}            & \acc{92.0}{0.24}  & \none{}           & \none{}          & \acc{15.9}{3.2}   & \none{}             & \none{}\\
GECA*           & \acc{87.0}{1.0}   & \none{}                & \none{}            & \acc{82.0}{4.0}   & \none{}           & \none{}          & \none{}           & \none{}             & \none{}\\
LANE            & \ACC{100.0}{}     & \none{}                & \none{}            & \ACC{100.0}{}     & \none{}           & \none{}          & \ACC{100.0}{}     & \ACC{100.0}{}    & \none{}\\
\hline
Meta seq2seq*    & \ACC{99.9}{}      & \none{}               & \none{}            & \ACC{99.9}{}      & \none{}           & \none{}          & \acc{16.64}{}    & \none{}             & \none{}\\
Synth*           & \ACC{100.0}{}    & \none{}                & \none{}            & \ACC{100.0}{}     & \none{}           & \none{}          & \ACC{100.0}{}    & \none{}             & \none{}\\
\hline
NSEN            & \gacc{0.0}{0.0}   & \gacc{0.0}{0.0}        & \gacc{0.0}{0.0}    & \gacc{0.0}{0.0}   & \gacc{0.0}{0.0}   & \gacc{0.0}{0.0}  & \gacc{0.0}{0.0}  & \gacc{1.7}{0.9} & \gacc{2.8}{0.3}\\
\hline
T5-small-NP     & \gacc{1.4}{0.8}    & \gacc{45.7}{15.4}     & \GACC{100.0}{0.0}  & \gacc{5.3}{4.6}   & \gacc{30.5}{8.7}  & \gacc{44.6}{11.2}& \gacc{19.4}{0.8}  & \gacc{0.9}{0.5} & \gacc{21.4}{1.5}\\
T5-small        & \gacc{84.1}{1.0}   & \gacc{73.0}{5.8}      & \GACC{100.0}{0.0}  & \gacc{31.8}{1.0}  & \gacc{58.2}{10.4} & \gacc{88.7}{8.9} & \gacc{10.9}{}     & \gacc{6.9}{1.1} & \gacc{28.0}{0.6}\\
T5-base         & \GACC{99.5}{0.0}   & \gacc{62.0}{0.9}      & \gacc{99.3}{0.3}   & \gacc{33.2}{0.5}   & \gacc{99.2}{0.2}  & \gacc{73.5}{1.8} & \gacc{14.4}{}     & \gacc{15.4}{1.1}& \gacc{31.2}{1.3}\\
T5-large        & \gacc{98.3}{}      & \gacc{69.2}{}         & \GACC{99.9}{}      & \gacc{46.8}{}     & \GACC{100.0}{}    & \gacc{91.0}{}    & \gacc{5.2}{}      & \gacc{10.1}{1.6}    & \gacc{34.8}{1.5}\\
T5-3B           & \gacc{99.0}{}      & \gacc{65.1}{}         & \GACC{100.0}{}     & \gacc{27.4}{}     & \gacc{90.0}{}   & \gacc{76.6}{}    & \gacc{3.3}{}      & \gacc{11.6}{}   & \gacc{40.2}{4.2}\\
T5-11B          & \gacc{98.3}{}      & \gacc{87.9}{}         & \GACC{100.0}{}     & \gacc{49.2}{}     & \gacc{99.1}{}     & \gacc{91.1}{}    & \gacc{2.0}{}      & \gacc{9.1}{}    & \gacc{40.9}{4.3}\\
T5-11B-mod      & \none{}            & \none{}               & \none{}            & \none{}            & \none{}           & \none{}          & \none{}           & \none{}          &\GACC{42.1}{9.1}\\
\end{tabular}
\setlength{\tabcolsep}{6pt}
}
\caption{Accuracy of various models (first column) on the traditional SCAN splits, the SCAN MCD-mean split (second-to-right column), and the CFQ MCD-mean split (rightmost column). Models in the first group are general-purpose architectures; the second group are SCAN-inspired approaches; the third group are SCAN-inspired approaches requiring a special meta-learning task setup; the fourth group are architectures designed for algorithmic learning; the last group are T5-based models. For the MCD splits, the reported variance is the 95\% confidence interval, while for all other splits it is the stdev. Models marked with * take additional knowledge as side inputs. Cells with a white background are results obtained in previous papers; cells with a grey background are results obtained in this paper. Boldfaced results are 0.5\% points within the best result.}
\label{table:results}
\end{table*}

Table~\ref{table:results} summarizes results on all evaluated models, together with results from prior work on SCAN and CFQ. Accuracy results with a white background are existing results, while those with a grey background are results we obtain in this paper. We provide additional experiments on the MCD splits for both SCAN and CFQ in Appendix~\ref{appendix:additional-experiments}. We make four main observations, which we describe in the next subsections.

\subsection{Masked Language Model Pre-training}
\emph{Pre-training helps for compositional generalization, but doesn't solve it.}\\

\noindent
Comparing T5-small-NP to T5-small, we see that pre-training helps significantly on CFQ, and an increase in T5 model size is correlated with an increase of accuracy on MCD. Furthermore, pre-training is beneficial for all SCAN splits involving holdouts of primitives, subcommands, or templates, with the biggest gains on the ``Add jump'' primitive holdout split. However, it leads to an average decrease in accuracy of 8.5\% on the length split. In Appendix~\ref{appendix:pretraining}, we show that on the CFQ MCD split, gains were limited to lengths that were seen in training.

We hypothesize that the primary benefit provided by pre-training is to improve the model's ability to substitute similar words or word phrases by ensuring they are close to each other in the representation space. The scenario which we would expect to benefit most from such a mechanism is that of substitution of a single-word primitive, typified in the ``Add jump'' task, where pre-training indeed succeeds in achieving near-perfect performance, with lesser gains on the splits requiring substitution of multi-word phrases, subcommands, or groups of phrases described by templates. Pre-training does not, however, fully solve the compositional generalization tasks. We would need further investigation to determine whether pre-training's negative effect on length generalization means that pre-training actively harms the model's ability to learn to build larger constructs through composition of known building blocks. 

Note that while the pre-trained T5 models consistently outperform T5-small-NP on splits other than the length split, relative performance is not fully consistent between Transformer and T5-small-NP, which are both (non-pre-trained) Transformers. These differences are unrelated to pre-training per se and, similarly to inconsistent relative performance described in Section~\ref{sect:general-seq-to-seq-results}, would require additional investigation to explain.

\subsection{Specialized Architectures}
\emph{For the specialized architectures we evaluated, improvements obtained on one compositional generalization benchmark do not transfer to others.}\\

\noindent
CGPS -- while showing strong performance on traditional SCAN splits -- performs poorly on MCD splits, and is outperformed by all general-purposes architectures on CFQ and by the basic LSTM+A architecture on SCAN. As shown in Appendix~\ref{appendix:additional-experiments}, the underperformance of CGPS vs. other general-purpose architectures is even more noticeable at moderate levels of compound divergence. While it would be unsurprising for CGPS to show only limited gains in high compound divergence splits due to the relatively small number of examples in those that can be directly solved by primitive substitution alone, this is insufficient to explain why CGPS performs worse than general-purpose architectures. It appears rather that the CGPS mechanism, unlike pre-training, is not robust to shifts in compound distribution and even introduces negative effects in such circumstances.

NSEN performs very poorly on all of the splits we evaluated. It may be noted that while we did manage to train NSEN to 98.3\% on a SCAN random split, on CFQ the maximum accuracy we were able to obtain is 56.6\% even on a random split. One noteworthy difference between NSEN and the other architectures is that NSEN is not autoregressive. However, in Appendix~\ref{appendix:ablation}, we show through an ablation study that autoregression is only a small factor in the stronger performance of LSTM+A, suggesting it is unlikely to fully explain the poor performance of NSEN. We hypothesize that while NSEN has been shown to perform well on algorithms and simple language tasks that align cleanly to its shuffle exchange architecture, it is insufficiently general to efficiently learn the more complex task of semantic parsing. This hypothesis is supported by NSEN's relatively poor performance on even the random split of the more structurally complex CFQ task.

A remaining question for follow-up research is whether the most recent specialized approaches LANE and Synth (which have been developed concurrently with this work) could break this trend, although we expect that both of them would have to be adapted substantially for CFQ.

\subsection{General Sequence-to-sequence Architectures}
\label{sect:general-seq-to-seq-results}
\emph{General ML architecture improvements yield incremental, if limited, improvements in compositional generalization settings.}\\

\noindent
On CFQ MCD, the performance order of models (Evolved Transformer $>$ Transformer $>$ LSTM+A) corresponds with what we would expect based on results obtained in other areas such as machine translation~\cite{vaswani2017attention,evolved_transformer}. On SCAN MCD, LSTM+A outperforms all other general architectures, but this seems to be a peculiarity not present in lower compound divergence DBCA splits (see Figure~\ref{fig:overall-results} in Appendix~\ref{appendix:additional-experiments}). On 5 out of 7 of the other SCAN splits, Transformer outperforms LSTM+A, while Evolved transformer outperforms Transformer.

Compared to the Transformer and the Universal Transformer, the Evolved Transformer's performance on the length split is much better, and mirrors a similar performance improvement in T5-small-NP, another Transformer variant. Further comparison between these architectures is needed to explain this surprising result.

\subsection{Interpretation of Prior Work}
\label{sect:interpreting-prior-work}
\emph{It is challenging to evaluate compositional generalization using the traditional SCAN splits.}\\

\noindent
While the SCAN splits have served an important purpose in motivating innovative research on compositional generalization, the number of splits is somewhat unwieldy, and perhaps as a result, many models have not been evaluated on all splits. Furthermore, as seen, for example, in comparison of performance between Transformer and Evolved Transformer on ``Around right'' vs. ``Opposite right'', or between Transformer and T5-small on ``Add jump'' vs. ``Add turn left'', the relative performance of models is not strongly correlated across splits.
This makes it difficult to perform an unambiguous evaluation of the compositional generalization abilities of different architectures. MCD splits alleviate this problem by providing a single yet comprehensive measure for compositional generalization.

\section{Related Work}
\label{sect:related-work}

We discuss here additional lines of research related to improving compositional generalization of ML systems and performance of semantic parsing systems in settings other than SCAN or CFQ. Many involve ideas potentially applicable to CFQ or SCAN and suggest potential areas of future work.

\paragraph{Pre-training on artificial languages}
While we show that MLM pre-training already yields significant performance improvements on both CFQ and SCAN, it has likely not achieved its full potential on these tasks, as we pre-train currently only on natural language -- the input language -- whereas the output (SPARQL or SCAN actions) is in an artificial language with a distinct syntax, and whose tokens have at most a loose connection to English words. Recent research has shown success in applying language model pre-training to artificial languages as well, which could potentially be applicable to CFQ, given a sufficiently large SPARQL corpus. Specifically, \citet{lachaux2020unsupervised} apply unsupervised machine translation techniques to the problem of source-to-source compilation, achieving competitive performance by training a language model on documents from each of three programming languages, while relying on tokens with common meaning across these languages to achieve a joint embedding space. \citet{feng2020codebert} treat natural language (NL) and programming language (PL) as distinct modes and use a combination of paired examples and monolingual documents to train a bi-modal pre-trained model, which they apply to natural language code search and code documentation generation (i.e., PL-to-NL translation). The semantic parsing task of CFQ would involve similar requirements as PL-to-NL translation, but with the direction of translation reversed, while ideally using techniques of \citet{lachaux2020unsupervised} to depend only on monolingual sources.

\paragraph{Intermediate representations}
Intermediate representations to simplify the task of mapping input to output have been applied to a text-to-SQL task in the form of SemQL~\citep{guo2019complex}, which provides a simplified output space at the cost of some reduced expressivity, and to earlier text-to-SPARQL tasks using representations such as lambda-DCS~\citep{liang2013lambda,berant2013semantic}. Based on the improvements yielded by the basic intermediate representation adopted in this paper, we see potential for further benefit from development of improved intermediate representations that maintain precision and expressive power while supporting programmatic manipulation and closer mapping to natural language structures.

\paragraph{Data augmentation} 
Additional evidence in favor of GECA-like data augmentation is provided by \citet{hill2019environmental}, who show that implicit data augmentation effects in a situated agent task can improve compositional generalization. They also observe that generalization to previously unseen action combinations improves as the absolute number of primitive actions and objects observed in training increases, potentially motivating data augmentation techniques that combine known grammatical patterns with additional (possibly artificially generated) primitives. \citet{kagithasystematic} reports preliminary results suggesting that such a technique could improve performance on some SCAN splits. While the tendency to introduce incorrect examples is likely to remain a challenge, data augmentation has the advantage of architecture-independence and as future work could be combined with other approaches evaluated in this paper.

\paragraph{Syntax-constrained decoders}
In semantic parsing research motivated by the text-to-SQL datasets Spider~\citep{yu2018spider} and WikiSQL~\citep{zhong2017seq2sql}, a common technique is to use a syntax-constrained decoder to reduce the effective size of the output space and avoid errors in which the system produces malformed output. State-of-the-art systems on these two datasets constrain output via either a grammar-based decoder~\citep{wang2019rat} or a sketch-based decoder~\citep{choi2020ryansql}. Grammar-based decoders~\citep{yin2017syntactic} typically involve outputting a leftmost derivation sequence of the output parse tree, with the main gains coming from the relative ease of integration of syntax-based pruning~\citep{xiao2016sequence}. Sketch-based decoders~\citep{xu2017sqlnet} constrain output by forcing the output to follow a given high-level format called a ``sketch'', containing ``holes'' that the learned model must fill in. Both approaches are potentially applicable to the text-to-SPARQL task of CFQ, given the strict grammatical structure of SPARQL and, based on the results shown in this paper from simplification of output format, are likely to provide at least marginal improvements in performance.

\paragraph{Order independence in decoder}
Tasks like CFQ that output SPARQL or SQL have the potential to suffer from the ``order matters'' problem~\citep{vinyals2015order}, due to the mismatch between the sequential form of the expected output and the inherent order invariance of SPARQL constraints and SQL WHERE clauses. Early work on the WikiSQL dataset sought to mitigate this issue through integration of a sequence-to-set architecture~\citep{xu2017sqlnet} or through use of a non-deterministic oracle~\citep{shi2018incsql} to allow for multiple possible correct outputs, leading to modest gains in performance. Current state-of-the-art approaches for WikiSQL or Spider, however, do not adopt such mechanisms.

\paragraph{DB schema encoders}
Another standard technique on the Spider dataset is the use of graph encoders~\citep{li2015gated,gilmer2017neural} to encode the schema of the target database into a representation that can be attended to to influence both the encoding of the question and the decoding of the final output~\citep{shaw2019generating,bogin2019representing,bogin2019global,wang2019rat}. This technique is more directly relevant to Spider due to its specialized task setup, which focuses on generalization to databases unseen in training and thus depends on the database schema being passed as an additional input. We do believe, however, that encoding DB schema as explicit knowledge could potentially benefit compositional generalization by helping to disentangle the generic language understanding algorithm from the knowledge that should parameterize it and could thus be a relevant future area to investigate for tasks such as CFQ as well.

\paragraph{Other approaches}
Appendix~\ref{appendix:ext-related-work} describes approaches that are of interest from the broader perspective of compositional generalization, but are not directly applicable to the SCAN or CFQ tasks.

\section{Conclusion and Outlook}
We investigate state-of-the-art techniques and architectures to assess their effectiveness in improving compositional generalization in semantic parsing tasks based on the SCAN and CFQ datasets.

Our four main findings are as follows: First, pre-training helps for compositional generalization, but does not solve it. Secondly, for the specialized architectures we evaluated, improvements obtained on one compositional generalization benchmark do not transfer to others. Thirdly, improvements to general-purpose sequence-to-sequence architectures generally lead to corresponding incremental improvements in compositional settings. 
Fourthly and lastly, it is challenging to unambiguously and comprehensively evaluate compositional generalization using the traditional SCAN splits, which may be a reason for researchers to focus more on the MCD splits as a comprehensive measure of compositional generalization.

Our findings suggest several promising routes to further improve compositional generalization. As our pre-trained models were pre-trained on English (the input language) alone, further improvements amy be possible be pre-training on SPARQL (the output language) as well. Performance gains from simplification of output format suggest potential further gains could be achieved through adoption of improved intermediate representations and/or syntax-constrained decoding. We are interested in evaluating promising approaches such as LANE and Synth that would require more substantial adaptation to CFQ. We also welcome future work evaluating architectures and techniques from related domains on CFQ, as described in the related work.


\section{Acknowledgements}
We thank: Lukasz Kaiser for useful discussions and pointing us to NSEN; Yuanpeng Li, for interesting discussions and pointers to get CGPS working; Emīls Ozoliņš, for helping us set up NSEN; Adam Roberts for helping us debug the T5 pipeline; Pete Shaw, for pointing us to the intermediate representations that were used in text-to-SQL datasets; Olivier Bousquet for providing guidance and insights, and finally, we thank Xiao Wang for being involved in many of the discussions around our progress and providing useful insights.

\bibliography{references}

\clearpage
\appendix
\section*{Appendix}

\setcounter{topnumber}{5}
\setcounter{bottomnumber}{5}
\setcounter{totalnumber}{10}
\renewcommand{\topfraction}{0.9}
\renewcommand{\bottomfraction}{0.9}
\renewcommand{\textfraction}{0.1}
\renewcommand{\floatpagefraction}{0.9}

\section{Contributions}
\label{sect:contributions}

Daniel ran the experiments for the Evolved transformer, NSEN and CGPS. Marc set up the T5 pre-training infrastructure, performed the experiments, and developed the intermediate SPARQL representation. Nathan researched related work. Nathanael provided high-level project direction. All authors helped set the scope and research direction and contributed to the writing of the paper.

\section{Hyperparameters}
\label{suppl:hyperparameters}

We provide the hyperparameters used for our experiments in Table~\ref{tab:hyperparameters}.
All hyperparameters searches were done on a random split of the CFQ dataset.

\begin{table*}[ht!]
    \centering \small
    \begin{tabular}{@{}lccc@{}}
    \hline
        &  CGPS (CFQ) & Evolved Transformer & NSEN \\
        \hline \hline 
        train steps             & 8,000 & 100,000   & 100,000 \\
        batch size              & 2,048 & 4,096     & 4,096\\
        hidden size             & 512   & 128       & 384 \\
        embedding size          & 512   & --        & -- \\
        function embedding size & 256   & --        & -- \\
        num hidden layers       & --    & 2         & 2 \\
        num heads               & --    & 8         & -- \\
        learning rate schedule  & --    & constant*single\_cycle\_cos\_decay & noam \\
        learning rate \{,constant\}  & 0.0013319345874256959 & 0.0011703123695332683  & 0.1 \\
        learning rate warmup steps & -- & 4,000  & 100\\
        dropout                 & --    & -- & 0.1 \\
        \hline
    \end{tabular}
    \caption{Summary of hyperparameters that deviate from the defaults.}
    \label{tab:hyperparameters}
\end{table*}




\section{Ablation Study: Autoregression}
\label{appendix:ablation}

\begin{figure}[ht!]
    \centering
    \includegraphics[width=\columnwidth]{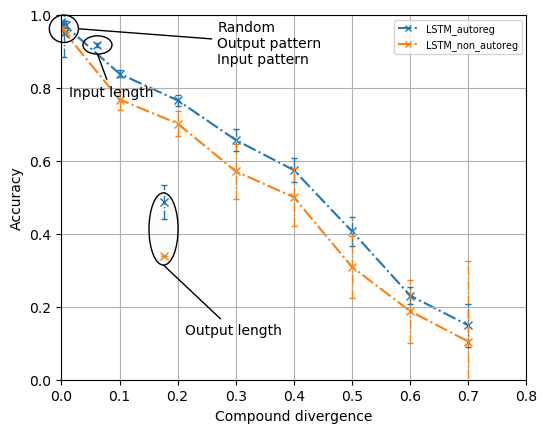}
    \caption{Autoregression ablation, LSTM+A on CFQ.}
    \label{fig:non-autoregressive}
\end{figure}

One significant difference between NSEN and other architectures is that the NSEN is not autoregressive. Autoregression has been found to be important for generative natural language tasks such as machine translation and the NACS (reverse SCAN) task~\citep{kaiser2016active, bastings2018jump}. We show that this is unlikely to explain why NSEN is performing so poorly, however, by running an ablation experiment where we remove autoregression from the LSTM baseline and observe only minimal degradation in performance (see Figure~\ref{fig:non-autoregressive}). It is also worth noting that~\citet{bastings2018jump} observes that removing autoregression improves performance on SCAN, which is then given as an argument that simpler architectures do better on SCAN. Our results show that this is not the case for CFQ.

\section{Additional Experiments on the MCD Splits}
\label{appendix:additional-experiments}

\begin{figure*}[ht!]
    \centering \footnotesize
    (a)\includegraphics[width=0.4\linewidth]{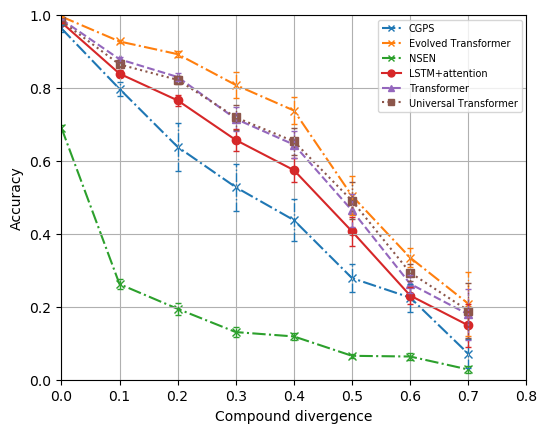}%
    \hfill%
    (b)\includegraphics[width=0.4\linewidth]{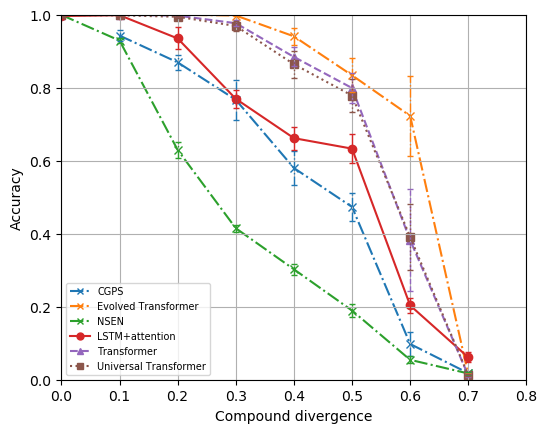}%
\caption{Accuracies of the various models on (a) CFQ and (b) SCAN{} vs.\ compound divergence for different split methods and for different target compound divergences.}
    \label{fig:overall-results}
\end{figure*}

We apply the DBCA method to evaluate CGPS and NSEN on both CFQ and SCAN, and compare them with the baselines of~\citet{Keysers2020}: LSTM+A, Transformer, and Universal Transformer (Figure~\ref{fig:overall-results}). Our observations are as follows:

The relative performance of architectures on the CFQ MCD splits are observed consistently across lower compound divergences, starting from even modest divergences of 0.1 to 0.2. 

Benefits of Evolved Transformer on SCAN are more noticeable at medium divergences. In particular, at all points other than the MCD, Evolved Transformer matches or outperforms LSTM+A and CGPS (as well as the other architectures), consistent with the improvement observed in the MCD split of CFQ.

NSEN performance drops particularly steeply (on CFQ even more than on SCAN), suggesting that this architecture is particularly brittle with respect to distribution shifts.

\begin{table}[ht!]
\centering
\setlength{\tabcolsep}{3pt}
\resizebox{\columnwidth}{!}{
\begin{tabular}{lrlrlrlrl}
             & \mc{2}{MCD} & & & \\
\textbf{Model}              & \mc{2}{Mean} & \mc{2}{MCD1} & \mc{2}{MCD2} & \mc{2}{MCD3}\\
\hline
\hline
LSTM+A                      & \acc{6.1}{1.7}     & \acc{4.7}{2.2}   & \acc{7.3}{2.1} & \acc{1.8}{0.7} \\
Transformer                 & \acc{0.9}{0.3}     & \acc{0.4}{0.4}   & \acc{1.8}{0.4} & \acc{0.5}{0.1} \\
Univ. Trans.                & \acc{1.1}{0.6}     & \acc{0.7}{1.0}   & \acc{1.5}{0.2} & \acc{1.0}{0.6} \\
Evol. Transf.               & \acc{1.6}{0.6}     & \acc{1.4}{0.2}   & \acc{2.7}{1.2} & \acc{0.7}{0.4} \\
\hline
CGPS                        & \acc{2.0}{0.7}     & \acc{2.6}{1.2}   & \acc{1.8}{0.6} & \acc{1.4}{0.5} \\
NSEN                        & \acc{1.7}{0.9}     & \acc{1.2}{1.0}   & \acc{1.2}{0.6} & \acc{2.0}{1.0} \\
\hline
T5-small-NP                 & \acc{0.9}{0.5}     & \acc{0.2}{0.3}   & \acc{1.7}{0.7} & \acc{0.9}{0.4}\\
T5-small                    & \acc{6.9}{1.1}     & \acc{9.5}{1.2}   & \acc{2.4}{0.9} & \acc{8.8}{1.1} \\
T5-base                     & \ACC{15.4}{1.1}    & \ACC{26.2}{1.7}  & \ACC{7.9}{1.6} & \acc{12.1}{0.1}\\
T5-large                    & \acc{10.1}{1.6}    & \acc{17.7}{3.4}  & \acc{2.3}{0.8} & \acc{10.3}{0.6}\\
T5-3B                       & \acc{11.6}{}       & \acc{22.2}{}     & \acc{3.5}{}    & \acc{9.2}{}\\
T5-11B                      & \acc{9.1}{}        & \acc{7.9}{}      & \acc{2.4}{}    & \ACC{16.8}{}\\
\end{tabular}
}
\setlength{\tabcolsep}{6pt}
\caption{Various models accuracy on SCAN MCD. The variance reported is the 95\% confidence interval.}
\label{table:results-mcd-scan}
\end{table}

\begin{table}[ht!]
\centering
\resizebox{\columnwidth}{!}{
\setlength{\tabcolsep}{3pt}
\begin{tabular}{lrlrlrlrl}
                & \mc{2}{MCD}        &  & & &  & \\
\textbf{Model}  & \mc{2}{Mean}     & \mc{2}{MCD1} & \mc{2}{MCD2} & \mc{2}{MCD3}\\

\hline
\hline
LSTM+A          & \acc{14.9}{1.2}  & \acc{28.9}{1.8} & \acc{5.0}{1.1}  & \acc{10.8}{0.6} \\
Transformer     & \acc{17.9}{0.8}  & \acc{34.9}{1.1} & \acc{8.2}{0.3}  & \acc{10.6}{1.1} \\
Univ. Trans.    & \acc{18.9}{1.4}  & \acc{37.4}{2.2} & \acc{8.1}{1.6}  & \acc{11.3}{0.3} \\
Evol. Trans.    & \acc{20.8}{0.7}  & \acc{42.4}{1.0} & \acc{9.3}{0.8}  & \acc{10.8}{0.2} \\
\hline
CGPS            & \acc{7.1}{1.8}   & \acc{13.2}{3.9} & \acc{1.6}{0.8}  & \acc{6.6}{0.6} \\
NSEN            & \acc{2.8}{0.3}   & \acc{5.1}{0.4}  & \acc{0.9}{0.1}  & \acc{2.3}{0.3} \\
\hline
T5-small-NP     & \acc{21.4}{1.5}  & \acc{42.5}{2.6} & \acc{11.2}{1.5} & \acc{10.6}{0.4} \\
T5-small        & \acc{28.0}{0.6}  & \acc{54.2}{0.8} & \acc{16.0}{0.3} & \acc{13.8}{0.8} \\
T5-base         & \acc{31.2}{1.3}  & \acc{57.6}{1.4} & \acc{19.5}{1.0} & \acc{16.6}{1.5} \\
T5-large        & \acc{34.8}{1.5}  & \acc{63.3}{0.6} & \acc{22.2}{1.5} & \acc{18.8}{2.6} \\
T5-3B           & \acc{40.2}{4.2}  & \ACC{64.0}{1.5} & \acc{29.7}{2.8} & \acc{27.0}{8.3} \\
T5-11B          & \acc{40.9}{4.3}  & \acc{61.4}{4.8} & \acc{30.1}{2.2} & \acc{31.2}{5.7} \\
T5-11B-mod      & \ACC{42.1}{9.1}  & \acc{61.6}{12.3} & \ACC{31.3}{12.8} & \ACC{33.3}{2.3} \\
\end{tabular}
\setlength{\tabcolsep}{6pt}
}
\caption{Various models accuracy on CFQ MCD. The variance reported is the 95\% confidence interval.}
\label{table:results-mcd-cfq}
\end{table}

We also provide full accuracy results of all MCD splits on both SCAN (Table~\ref{table:results-mcd-scan}) and CFQ (Table~\ref{table:results-mcd-cfq}) for the architecture that we evaluated in Figure~\ref{fig:overall-results}, as well as the full pre-training results. 

\section{Intermediate SPARQL Representation}
\label{appendix:sparql}

In this section we provide details on the intermediate representation used in the model T5-11B-mod.

Recall from Section~\ref{sect:output-simplification} that the question ``Did M0 and M1 direct M2 and M3?'' has the following corresponding SPARQL clauses.

\begin{quote}
\texttt{M0 directed M2 . M1 directed M2 . M0 directed M3 . M1 directed M3}
\end{quote}

Representing the clauses of a SPARQL query by $\{c_1,\ldots, c_n\}$, where $c_i=(subj_i,rel_i,obj_j)$ with $1\le i\le n,$ we build the intermediate representation by applying the following three transformations in sequence to the original clauses.

\paragraph{Group subjects ($f_1$)} Replace clauses $(subj, rel1, obj1),\ldots (subj, relk, objk)$ by $(subj, {(rel1, obj), \ldots, (relk, objk)})$. Applying this transformation to our example gives the following intermediate representation.

\begin{quote}
\texttt{M0 \{ directed M2 . directed M3 \} M1 \{ directed M2 . directed M3 \}}
\end{quote}

\paragraph{Group subjects and objects ($f_2$)} Apply $f_1$, and replace clauses $(subj, \{(rel, obj_1), \ldots,$ $(rel, obj_p), (rel_1, obj_{p+1}), \ldots, (rel_s, obj_{p+s})\})$ by $(subj, \{(rel, \{obj_1, \ldots,obj_p\}), (rel_1, \{obj_{p+1}\}), \ldots,$ $(rels, \{obj_{p+s}\})\})$. Applying this transformation to our example gives the following intermediate representation.

\begin{quote}
\texttt{M0 \{ directed \{ M2, M3 \} \}}\\
\texttt{M1 \{ directed \{ M2, M3 \} \}}
\end{quote}

\paragraph{Group subjects and objects and sort alphabetically ($f_3$)} The last representation is exactly the same as $f_2$, but ensuring that the clauses are sorted. (In this particular example, the clauses already happened to be in sorted order, so there would be no effect.)

\begin{figure}[ht!]
    \centering
    \includegraphics[width=0.8\columnwidth]{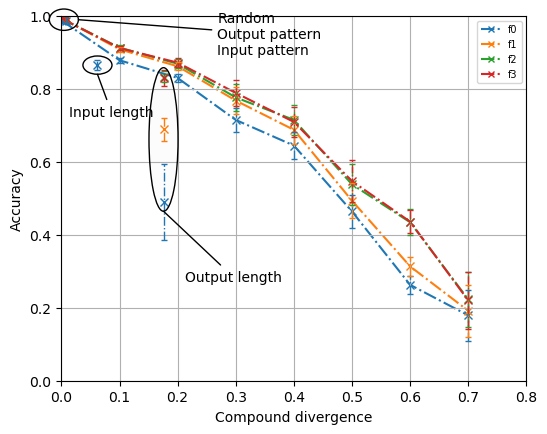}
    \caption{\footnotesize Transformer accuracy on CFQ with intermediate SPARQL representations (see Section~\ref{appendix:sparql}.}
    \label{fig:intermediate-sparql}
\end{figure}

Figure~\ref{fig:intermediate-sparql} shows the effect of using different intermediate representations on the compound divergence splits of CFQ for the Transformer. The results indicate that the intermediate representations lead to significant gains for all compound divergence splits. Grouping both subjects and objects leads to the biggest wins, and sorting alphabetically does not seem to affect accuracy significantly.

\section{Effect of Pre-training Input and Output Length}
\label{appendix:pretraining}

\begin{figure*}[ht!]
    \centering
    \includegraphics[width=\linewidth]{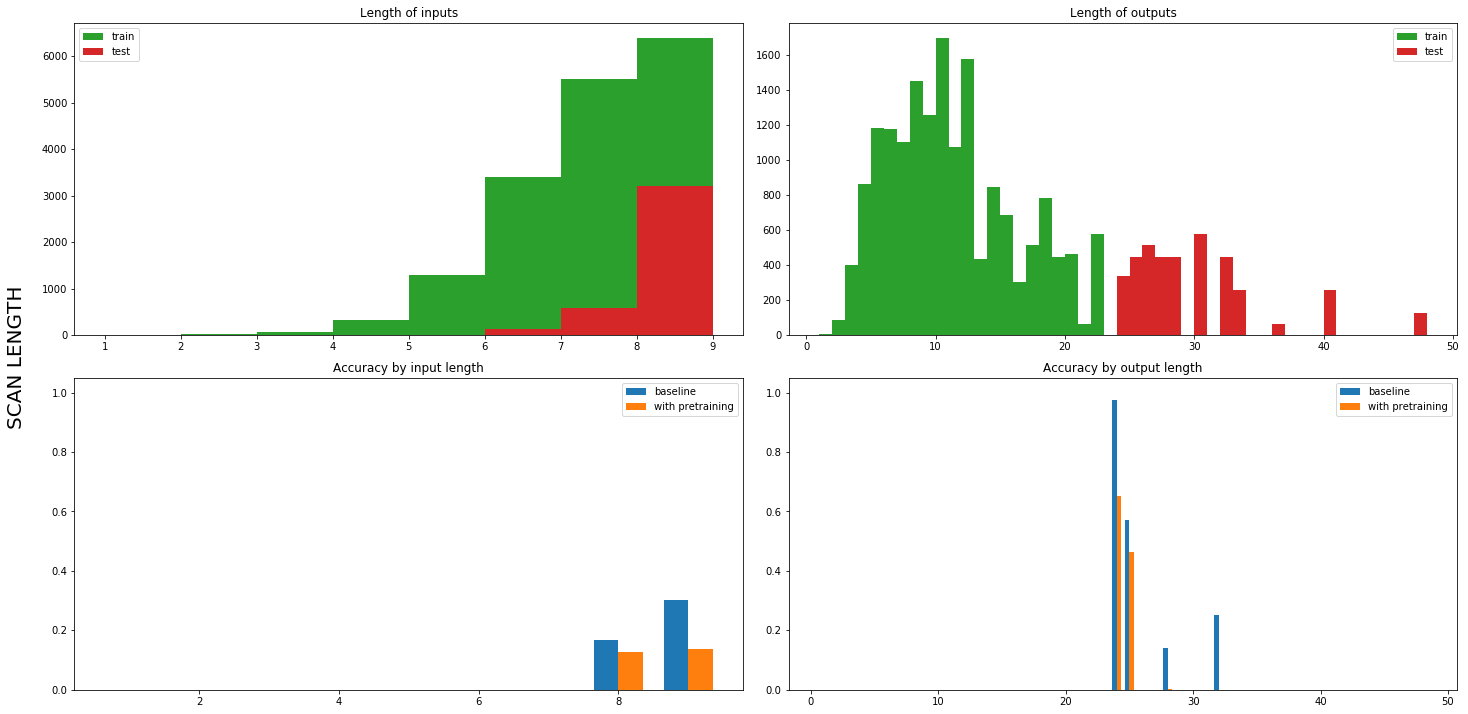}
    \includegraphics[width=\linewidth]{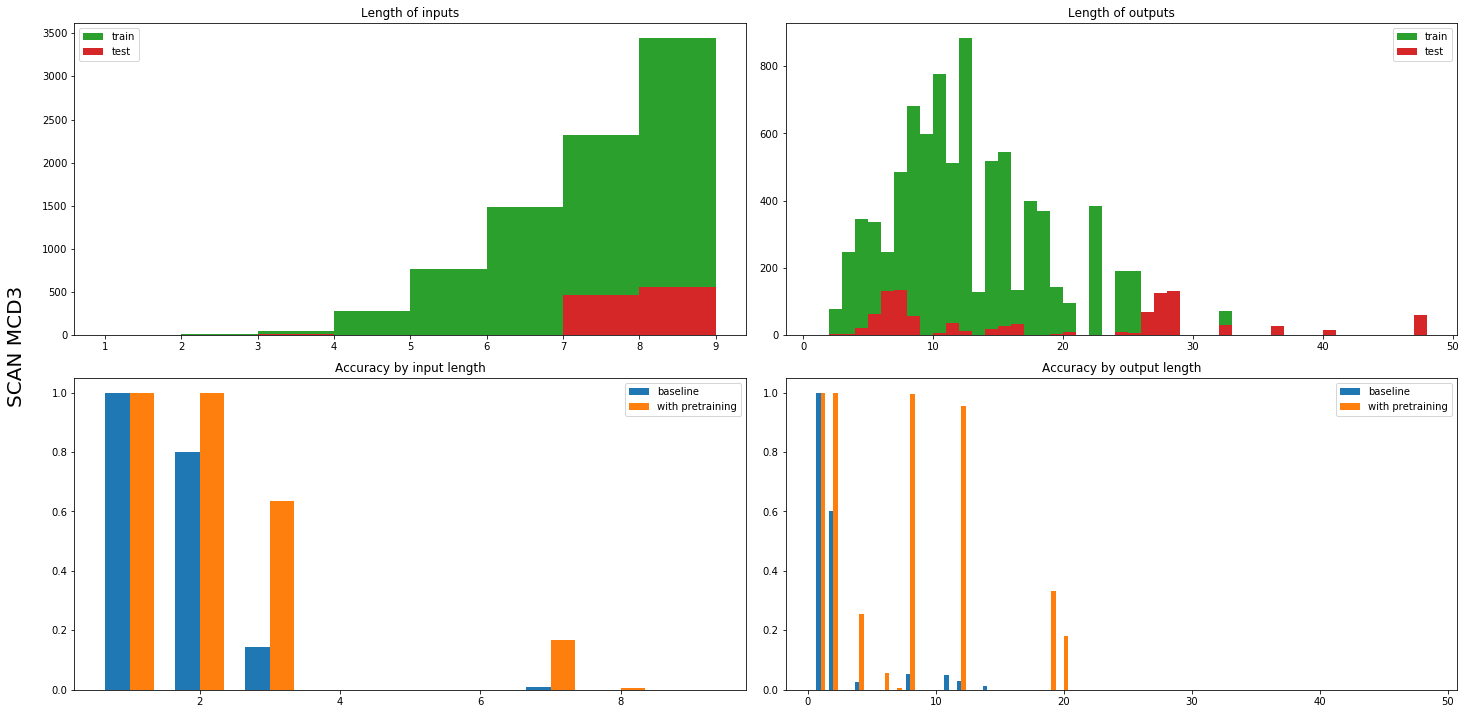}
    \includegraphics[width=\linewidth]{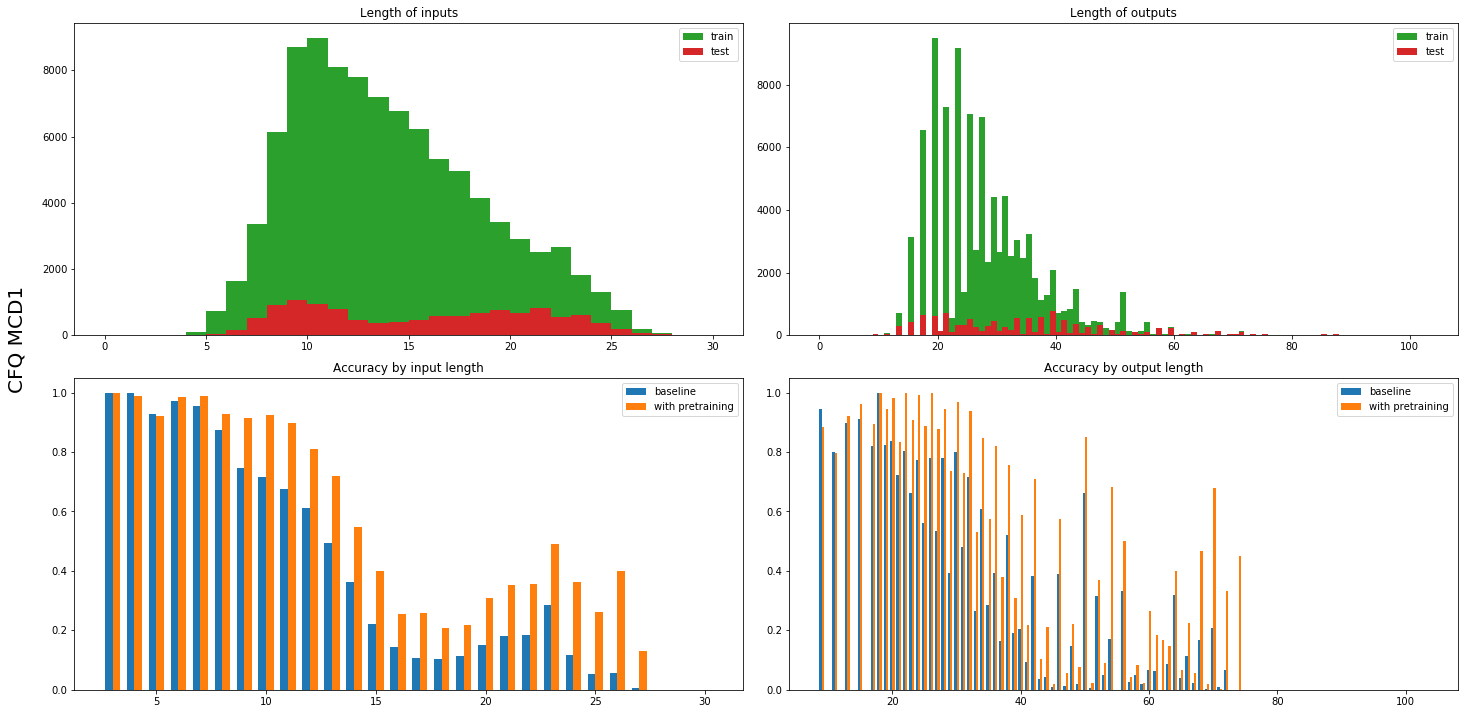}
    \caption{\footnotesize Length analysis of SCAN length, SCAN MCD3 and CFQ MCD1 splits. We show the distribution of train and test samples as well as the performance impact of pre-training broken down by length.}
    \label{fig:length-analysis}
\end{figure*}

We analyze the effect of pre-training further by comparing T5-small-NP (named the ``base'' model below) with T5-small (the ``pre-trained'' model).  While pre-training is beneficial for many splits, on the length split on SCAN it leads to an average decrease in accuracy of 8.5\%. Figure~\ref{fig:length-analysis} (top) shows a breakdown of the input and output lengths for this split. The test set of this split has input examples of length 8 or 9. As seen in the figure, the relative performance of the pre-trained model compared to the base model largely decreases as the input or output length increases. We provide a similar breakdown for SCAN-MCD3 and for CFQ-MCD1 (Figure~\ref{fig:length-analysis} middle and bottom). While the pre-trained model outperforms the base model on these two splits, we see that no gains were achieved at sentence lengths beyond the maximum seen in training, at which point accuracy was consistently 0 both with and without pre-training. 

\section{Extended Related Work}
\label{appendix:ext-related-work}

This section describes approaches that have been used to encourage compositional generalization or to drive general performance improvements in question answering tasks, but which are not directly applicable to the SCAN or CFQ tasks.

\paragraph{DB content encoders}
A more extreme approach than that of DB schema encoding is to encode the actual content of the database, so as to attend to some relevant portion of this content when building a representation of the meaning of the questions and when seeking for an answer. This approach is taken by PullNet~\citep{sun2019pullnet}, the current state of the art on ComplexWebQuestions~\citep{talmor18compwebq}, a benchmark for question answering based on information potentially from web documents in addition to from a knowledge base. Semantic parsing literature typically treats approaches that consider database content as a separate task from approaches that are blind to database content, with the latter approach in some cases considered preferable due to its applicability to situations in which access to DB content is restricted for privacy reasons~\citep{zhong2017seq2sql,xu2017sqlnet,yu2018typesql}.


\paragraph{Neural module networks}
A number of benchmarks have been developed with the aim of testing some manner of compositional generalization in the area of visual question answering~\citep{CLEVR,bahdanau2019closure,bahdanau2019systematic,hudson2019gqa} and reading comprehension~\citep{dua2019drop}. A common approach in both of these spaces is the use of neural module networks~\citep{andreas2016neural,johnson2017inferring,bahdanau2019systematic,bahdanau2019closure,gupta2019neural}, which seek to encourage compositional generalization by decomposing the task into re-usable modules which can be combined at evaluation time in ways potentially unseen during training. In its current form, however, this approach depends as one of its components on a semantic parser for mapping the natural language question to a program layout, so is not an alternative for the semantic parsing task itself. In a related line of work, the Neural State Machine~\citep{hudson2019nsm} also uses a variation of a semantic parser internally to map natural language questions to sequences of reasoning instructions. In both of these cases, the innovations for providing a compositional inductive bias apply primarily to the mechanism by which the answer is retrieved from the relevant data source, rather than to the task of understanding the natural language question itself.

\end{document}